\documentclass[lettersize,journal]{IEEEtran}
\usepackage{amsmath,amsfonts}
\usepackage{algorithmic}
\usepackage{array}
\usepackage[caption=false,font=normalsize,labelfont=sf,textfont=sf]{subfig}
\usepackage{textcomp}
\usepackage{stfloats}
\usepackage{url}
\usepackage{verbatim}
\usepackage{graphicx}
\def\BibTeX{{\rm B\kern-.05em{\sc i\kern-.025em b}\kern-.08em
    T\kern-.1667em\lower.7ex\hbox{E}\kern-.125emX}}
\usepackage{balance}
\usepackage{multi row}
\usepackage{color,soul}

\begin{document}
\title{Learning From High-Dimensional Cyber-Physical Data Streams for Diagnosing Faults in Smart Grids}
\author{Hossein~Hassani,~\IEEEmembership{Graduate Student~Member,~IEEE},
        Ehsan~Hallaji,~\IEEEmembership{Graduate Student~Member,~IEEE},
        Roozbeh~Razavi-Far,~\IEEEmembership{Senior~Member,~IEEE},
        Mehrdad~Saif,~\IEEEmembership{Senior~Member,~IEEE},
\thanks{Hossein Hassani, Ehsan Hallaji, and Mehrdad Saif are with the Department of Electrical and Computer Engineering, University of Windsor, Windsor, ON N9B 3P4, Canada (e-mail: hassa12t@uwindsor.ca, hallaji@uwindsor.ca, msaif@uwindsor.ca).}
\thanks{Roozbeh Razavi-Far is with the Faculty of Computer Science and Canadian Institute for Cybersecurity, University of New Brunswick, Fredericton, NB E3B 5A3, Canada, and also with the Department of Electrical and Computer Engineering, University of Windsor, Windsor, ON N9B 3P4, Canada (e-mail: roozbeh.razavi-far@unb.ca).}
}

\markboth{}%
{How to Use the IEEEtran \LaTeX \ Templates}

\maketitle

\begin{abstract}
The performance of fault diagnosis systems are highly affected by data quality in cyber-physical power systems. These systems generate massive amounts of data that overburden the system with excessive computational costs. Another issue is the presence of noise in recorded measurements, which prevents building a precise decision model. Furthermore, the diagnostic model is often provided with a mixture of redundant measurements that may deviate it from learning normal and fault distributions. This paper presents the effect of feature engineering on mitigating the aforementioned challenges in cyber-physical systems. Feature selection and dimensionality reduction methods are combined with decision models to simulate data-driven fault diagnosis in a 118-bus power system. A comparative study is enabled accordingly to compare several advanced techniques in both domains. Dimensionality reduction and feature selection methods are compared both jointly and separately. Finally, experiments are concluded, and a setting is suggested that enhances data quality for fault diagnosis.
\end{abstract}


\section{Introduction}\label{sec1}
\IEEEPARstart{F}{ault} diagnosis system (FDS) is a critical component of cyber-physical power systems that is crucial for detecting malfunctions, identifying their cause, and pinpointing their location in the grid \cite{7069265}. Nevertheless, designing a traditional model-based FDS model becomes more challenging as the grid grows in scale and complexity. The integration of intelligent models with power systems results in a cyber-physical system that bypasses the aforementioned problem \cite{ZHANG2022107744}. Making use of data-driven approaches for diagnostic purposes provides the model with a better generalization. Furthermore, intelligent models facilitate the decision-making process in control systems by reducing human-machine interaction \cite{9525180}.

Data-driven FDS is primarily carried out using classification algorithms \cite{9241783, 8233109, 9548085}. Having a set of sampled records corresponding to different system states, one can build a statistical model that maps incoming signal patterns to certain conditions such as different events of faults in the systems \cite{hassani2021generative}. In a fault detection setting, the FDS model can be formulated as an anomaly detector that reveals any signal pattern mismatches those of the normal or healthy system state. The past decade has witnessed numerous advancements in intelligent FDS models that resort to sensory data for monitoring power systems \cite{8714012, 8970332}. Nevertheless, the computational efficiency of intelligent models and their dependency on data quality limit the feasibility of such methods for FDS in power systems.

Data-driven methods only perform well if supplied with high-quality data. These methods try to capture the characteristics of a data distribution based on the existing relationships in the input space and form a mapping from distribution samples to a set of states in the system \cite{9928313}. However, if the input distribution at hand is not a good representative of the system, the statistical model cannot capture the system characteristics accordingly. In power systems, this situation happens due to a number of reasons. Firstly, the captured signal through sensors often contains noise, which makes it harder for machine learning models to capture the intrinsic relationships within data. Furthermore, the abundance of the set of measurements and the number of sensors (e.g., phasor measurement units) result in high-dimensional data \cite{8999221}. The higher the dimensionality size of the data, the more samples are needed for obtaining an accurate statistical model. This phenomenon is also referred to as the curse of dimensionality in the literature. In turn, processing high-dimensional data significantly increases the computational burden of the FDS. In addition, the utilized measurements in the power system most likely result in some invariant and duplicate features that feed the FDS model with irrelevant information that deteriorates the precision of the constructed model.

Feature selection (FS) and dimensionality reduction (DR) techniques are two main approaches that are commonly used in tackling the curse of dimensionality and improving the data quality to ensure optimal performance of the diagnostic model. FS is referred to the process of filtering or ranking different features (i.e., dimensions of the feature space) to remove non-informative features and select only a limited set of features that lead to the optimal performance of decision-making models \cite{HASSANI2021104150, 7887916, 7410835}. On the other hand, DR methods transform the whole feature space into a smaller space \cite{PCA, LLE, 10.1145/3178155}. Under a supervised setting, DR may additionally improve distribution by making classes more distinguishable in the transformed space \cite{9609642, FisherLDA}.

In this paper, an experimental review is performed on state-of-the-art FS and DR methods for diagnosing faults in cyber-physical power systems. Both mentioned harsh conditions, namely noisy signals (with different levels and frequencies) and high-dimensional data are taken into account throughout the experiments. This comparative study is enabled by analyzing FS and DR methods in three study groups: 1) FS, 2) DR, and 3) FS and DR. Finally, we suggest the best methods that will most likely benefit FDS in real-world power systems, where similar conditions are expected.

The remainder of this paper is organized as follows. Section \ref{sec2} presents the employed case-study and data. Section \ref{sec3} explains the proposed methodology. Section \ref{sec4} contains the experimental results and analysis. Finally, the paper is concluded in Section \ref{sec5}.

\begin{figure*}
\centering
\includegraphics[width=0.75\textwidth]{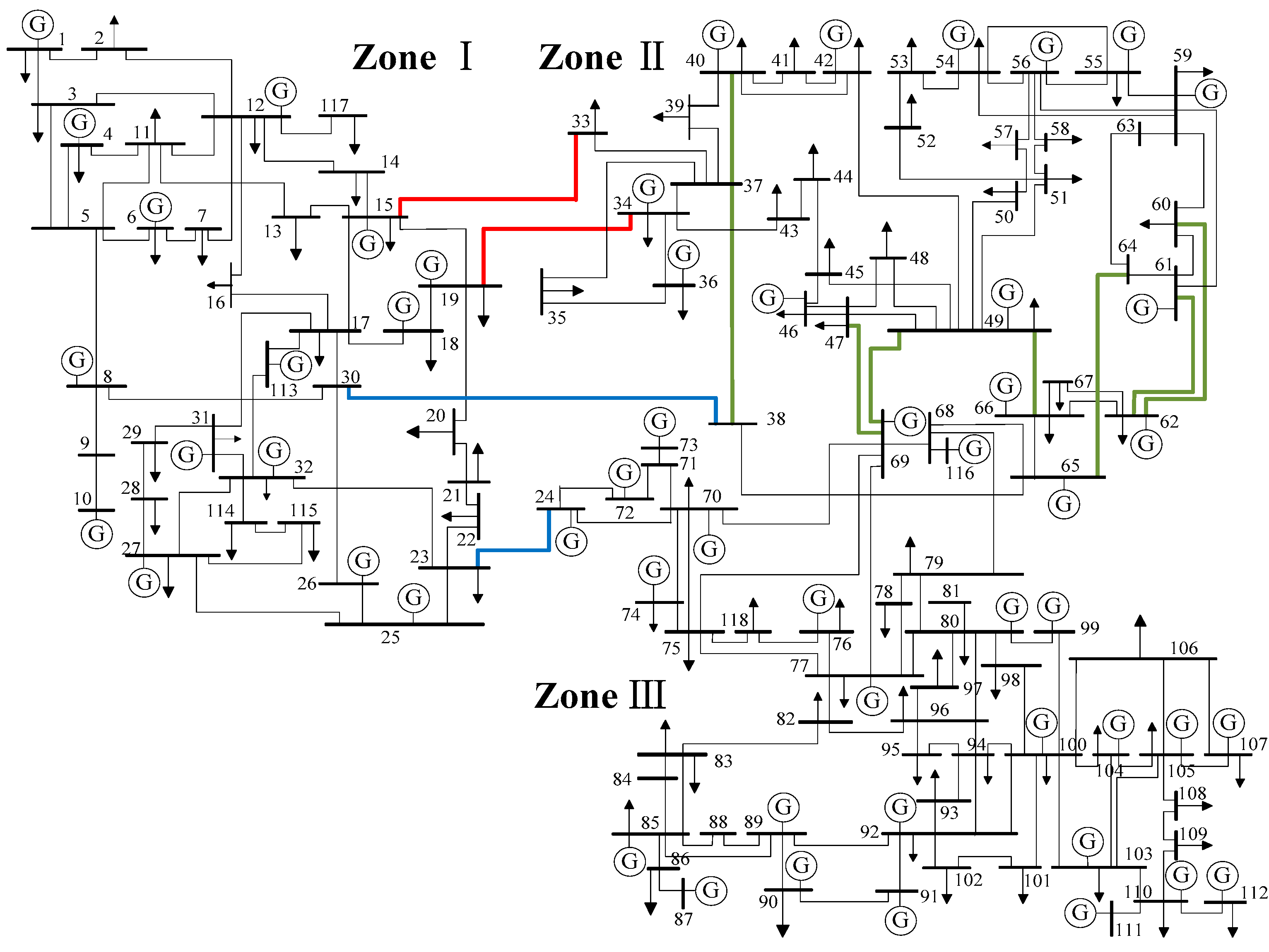}
\caption{The single-line diagram of the IEEE 118-bus system \cite{li2017robust}.}\label{fig21}
\end{figure*}

\section{Case-Study: Cyber-Physical Power Systems}\label{sec2}
In the present study, the IEEE 118-bus system has been employed in order to check for the effectiveness of the proposed diagnostic methodology. The single-line diagram of this system has been illustrated in Fig. \ref{fig21}. This system contains, as the name stands, 118 buses, 91 loads and 19 generation units. The system is simulated using PowerFactory, and data measurements are collected using a sampling rate of 10KHz for faulty and normal scenarios.

Three types of faults are simulated. These faults are named load-loss (LL), generator outage (GO), and generator ground (GG) faults. To simulate an LL or GO fault, a breaker has been placed between the load or generator and the corresponding bus. The breaker is initially closed, and, then, it is triggered for 25 ms to disconnect the load or generator from the bus. As for the GG faults, three-phase short-circuit faults are simulated between the generation units and the ground. In either fault case, data measurements are collected from the moment the fault appears in the system until the moment the fault has been cleared. This period of time has been set to be 25 ms and with a sampling rate of 10KHz, therefore, a total number of 250 samples are collected for each scenario. 

The list of simulated faults is represented in Table \ref{tab21}. As it can be observed from this table, there exist 31 LL fault scenarios, 19 GO, and 19 GG, in addition to the normal operational state of the system, leading to a total number of 70 simulated scenarios, where each one could be thought of as a class of data to be classified. Therefore, we are dealing with a multi-class classification problem with 70 classes. It is worth noting that as there are 19 generation units, the GO and GG faults are simulated on all of the buses connected to a generation unit. However, for the LL faults, only 31 out of 91 possible locations are considered in the construction of the fault scenarios. The selected buses to simulate the LL faults are tried to be from different zones of the grid as shown in Fig. \ref{fig21}, in order to consider the impact of the fault location on the performance of the proposed methodology. 

In general, six datasets are constructed following the aforementioned fault scenarios. These datasets are shown by $\{\mathcal{D}_1,\ldots,\mathcal{D}_6\}$ and the characteristics of each one is summarized in Table \ref{tab22}. We aim to investigate the effect of the signal-to-noise-ratio (SNR) and fault resistance (FR) values on the performance of the proposed diagnostic, too. In this regard, three different SNR values including $\{\text{10 dB}, \text{30 dB}, \text{70 dB}\}$ are considered to model the deep noisy measurements as well as data measurements with a slight noise. Following this, the FR value has also been set to be either $1\Omega$ or $10\Omega$. Therefore, with three SNR and two FR values, there could be a total of six combinations to construct the set of data, as described in Table \ref{tab22}. 

\begin{table}
\setlength{\tabcolsep}{35pt}
\centering
\caption{Summary of the simulated fault scenarios.}\label{tab21}
\begin{tabular}{ll}
\hline\hline Fault Type & \# Scenarios\\
\hline LL & 31\\
GO & 19\\
GG & 19\\
Normal & 1\\
\hline Total & 70\\
\hline\hline
\end{tabular}
\end{table}  

\begin{table}
\centering
\caption{Summary of the characteristics of the constructed datasets.}\label{tab22}
\begin{tabular}{ccccc}
\hline\hline Dataset & SNR (dB) & FR ($\Omega$) & \# Samples & \# Features\\
\hline $\mathcal{D}_1$ & 10 & 1 & 17,500 & 354\\
$\mathcal{D}_2$ & 10 & 10 & 17,500 & 354\\
$\mathcal{D}_3$ & 30 & 1 & 17,500 & 354\\
$\mathcal{D}_4$ & 30 & 10 & 17,500 & 354\\
$\mathcal{D}_5$ & 70 & 1 & 17,500 & 354\\
$\mathcal{D}_6$ & 70 & 10 & 17,500 & 354\\
\hline\hline
\end{tabular}
\end{table}

In terms of the set of features, as given in Table \ref{tab22}, there are a total number of 354 features. Firstly, three types of measurements are collected from each bus of the system following a simulated scenario. These measurements are voltage, frequency, and phase angle. Due to the fact that the system has 118 buses, and three types of features are collected from each bus, therefore, the constructed datasets contain 354 features. Table \ref{tab23} summarizes the set of features. 

\begin{table}
\centering
\setlength{\tabcolsep}{35pt}
\caption{Collected features to construct datasets.}\label{tab23}
\begin{tabular}{ll}
\hline\hline Measurements & Feature \\
\hline voltage & 1 to 118\\
frequency & 119 to 236\\
phase angle & 236 to 354\\
\hline\hline
\end{tabular}
\end{table}

\section{Methodology}\label{sec3}
\begin{figure}
    \centering
    \includegraphics[width=0.75\columnwidth]{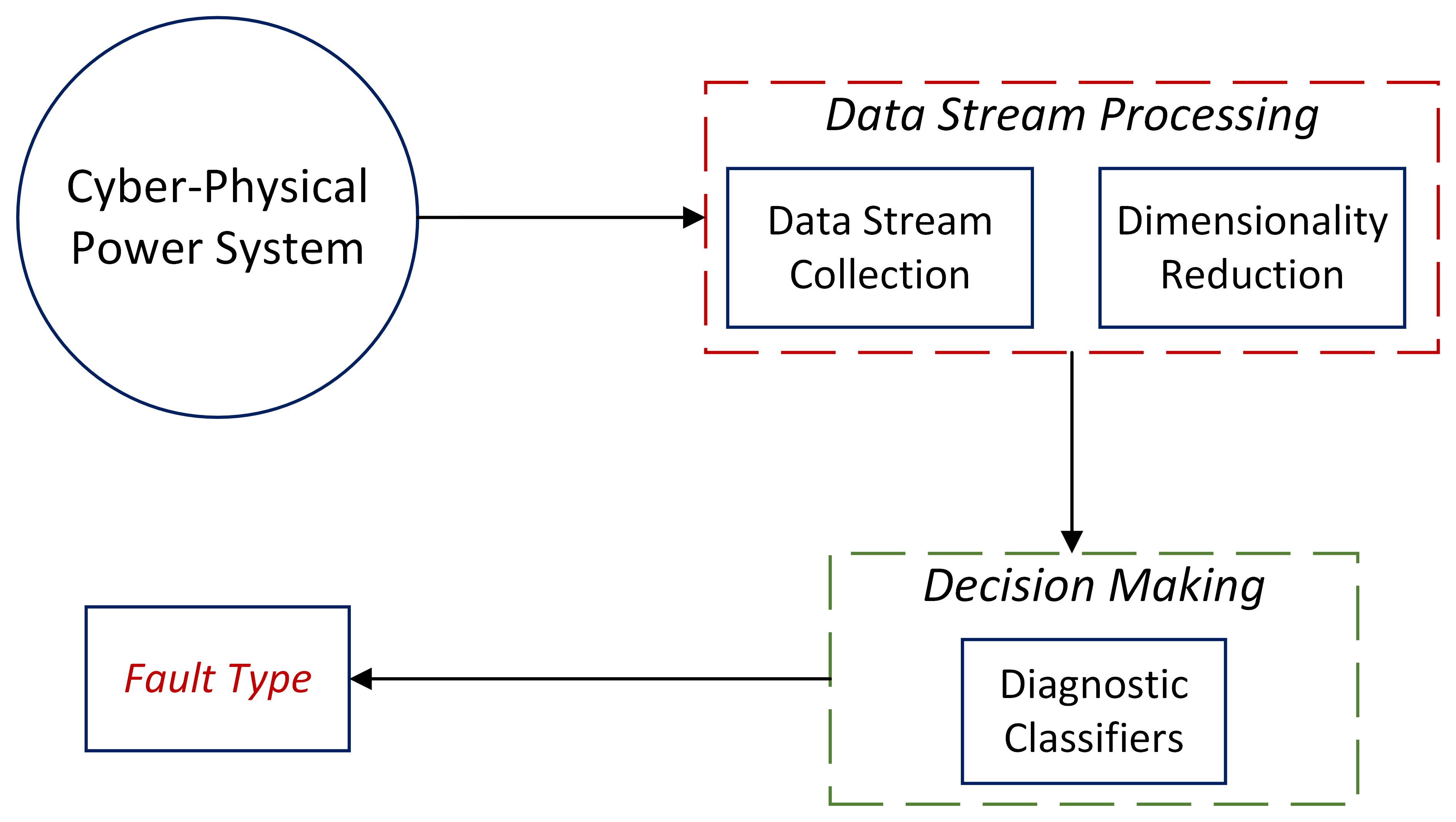}
    \caption{General framework of the proposed methodology.}
    \label{fig31}
\end{figure}

In line with what was previously mentioned in Section \ref{sec1}, the general framework of the proposed diagnostic model, as shown in Fig. \ref{fig31}, is to perform FS and DR techniques on the constructed datasets described in Table \ref{tab22}, and, then, fed the set of selected or extracted features to three classification models including k Nearest Neighbours (kNN), Support Vector Machine (SVM) and Random Forest (RF) for the sake of fault diagnosis. Within this general framework, we discuss the implementation procedure and evaluation metrics in this section. 

\subsection{Implementation Procedure}\label{sec31}
The first step in the implementation procedure is to standardize the given datasets. This is of paramount importance because the feature sets (i.e., voltage, frequency, and phase angle) take different values from different domains. For instance, the values of voltage measurements are per unit and are close to 1, while the frequency measurements fluctuate around 50 Hz. Therefore, there is a need to standardize the given datasets for the sake of eliminating the impact of scale variation from one feature to another. In this regard, given dataset $\mathcal{D}$, we do normalize the dataset column-wise through the following equation:
\begin{align}
\mathcal{D}^j_n = \frac{\mathcal{D}^j-\mathcal{D}^j_{min}}{\mathcal{D}^j_{max}-\mathcal{D}^j_{min}},
\end{align}
where $\mathcal{D}^j_n$ is the $j$th column of the normalized dataset with $j=1,\ldots,354$, $\mathcal{D}^j_{max}$ is the maximum value of a given column, and $\mathcal{D}^j_{min}$ shows the minimum value of the corresponding column. 

The normalized dataset will then be fed into the FS and DR modules in order to select or extract the set of most informative features. As for the FS techniques, we resort to seven techniques including Infinite Feature Selection (InfFS) \cite{7410835}, relief \cite{relief}, Least Absolute Shrinkage and Selection Operator (LASSO) \cite{7887916}, Unsupervised Feature Selection with Ordinal Locality (UFSOL) \cite{8019357}, and Concrete Feature selection based on Mutual Information (CFMI) \cite{7887916}. These techniques are selected to serve two purposes. First, the selected FS methods are categorized into filters, wrappers, and embedded. To this end, we aim to check which category of FS methods could improve diagnostic performance through a comparative study between different categories. Second, we aim to look for providing a ranking for the best FS-classifier combination for the sake of diagnosis. In the same vein, the selected DR techniques are from two broad categories namely linear and nonlinear, depending on the type of transformation. The employed DR techniques are Principal Component Analysis (PCA) \cite{PCA}, Linear Discriminant Analysis (LDA) \cite{FisherLDA}, Multi-dimensional Scaling (MDS) \cite{10.1145/3178155}, Locally Linear Embedding (LLE) \cite{LLE} and Constrained Adversarial Dimensionality Reduction (CADR) \cite{9609642}. The aforementioned two purposes are also considered in the implementation of DR techniques. Features of the selected FS and DR techniques are summarized in Table \ref{tab31}.

\begin{table}
\centering
\setlength{\tabcolsep}{15pt}
\caption{Characteristics of the selected FS and DR techniques.}\label{tab31}
\begin{tabular}{llll}
\hline\hline Technique & FS/DR & Type & Class\\
\hline InfFS & FS & Filter & Unsupervised \\
relief & FS & Filter & Supervised\\
UFSOL & FS & Wrapper & Unsupervised \\
LASSO & FS & Embedded & Supervised \\
CFMI & FS & Embedded & Unsupervised \\
PCA & DR & Linear & Unsupervised \\
LDA & DR & Linear & Supervised \\
MDS & DR & Nonlinear & Unsupervised \\
LLE & DR & Nonlinear & Unsupervised\\
CADR & DR & Nonlinear & Supervised \\
\hline\hline
\end{tabular}
\end{table}

Each FS or DR technique is then combined with a classification model (kNN, SVM and RF). In this regard, the selected set of features is fed into the aforementioned classification models and the classification results are then represented in terms of the accuracy and f-measure. However, the number of selected or extracted features by means of the FS and DR techniques needs to be adjusted carefully. To this end, in all experiments, we start with one feature and increase the number of features to the value, for which the performance of the classification model does not improve significantly. 

\subsection{Evaluation Metrics}\label{sec32}
To measure the performance of the FS/DR-classifier combinations, we resort to the classification accuracy $\mathcal{A}$ and f-measure $\mathcal{F}_1$ values. In this regard, the constructed combinations are validated through a 10-fold cross-validation manner and the attained results are collected in terms of accuracy and f-measure. It is evident that for binary classification with only positive and negative classes, the accuracy index captures the ratio of the correct decisions made by the classification model. As for the f-measure, it could be thought of as the harmonic mean of the precision $\mathcal{P}$ and recall $\mathcal{R}$, where the precision is a measure of the ratio of positive samples that are correctly classified into a positive category, while the recall measures the performance of the classification model in classifying the positive data samples. Following this, one case defines f-measure as given below:
\begin{align}\label{eq2}
\mathcal{F}_1=\frac{2\times \mathcal{P}\times\mathcal{R}}{\mathcal{P}+\mathcal{R}}.
\end{align}
In terms of the multi-class classification, however, a confusion matrix of the following form could be constructed:
\begin{align}
\mathcal{C}=\left[\begin{array}{ccc}cm_{11} & \ldots & cm_{1l}\\ \vdots & \ddots & \vdots \\ cm_{l1} & \vdots & cm_{ll}\end{array}\right],
\end{align}
With $l$ being the total number of classes. Following this structure, the precision and recall can then be defined as follows:
\begin{align}
\mathcal{P} & = \frac{1}{l}\sum_{k=1}^{l}\frac{cm_{kk}}{\sum_{i=1}^lc_{ki}},\\
\mathcal{R} & = \frac{1}{l}\sum_{k=1}^{l}\frac{cm_{kk}}{\sum_{i=1}^lc_{ik}}.
\end{align}
Having $\mathcal{P}$ and $\mathcal{R}$ calculated, the f-measure can then be obtained based on the Eq. \ref{eq2} for a multi-class classification problem. 

\section{Simulation Results and Comparative Study}\label{sec4}
In this section, a comprehensive comparative study is provided to compare the performance of FS/DR-classifier combinations. In this regard, the ultimate goal is to identify what combinations perform better for the sake of fault diagnosis in cyber-physical power systems. In terms of the FS and DR techniques, we initially investigate that what category of methods could deal better with the constructed datasets. In particular, FS techniques are selected from three categories including filters, wrappers, and embedded methods, and the selected DR techniques are either linear or nonlinear. Therefore, the goal is to investigate which category of FS and DR methods is superior in comparison with other categories. Furthermore, by resorting to the overall performance of FS and DR techniques, we aim to investigate whether FS techniques are preferred over DR techniques or vice versa. 

\subsection{Baseline}\label{sec41}
The baseline models refer to the case, in which no data reduction technique is employed. In this regard, the original datasets are only normalized and, then, fed into the classification models. Following this setup, the classification results are collected in terms of accuracy and f-measure through a 10-fold cross-validation scheme. To this end, the attained average accuracy and f-measure values by means of the baseline models are represented in Table \ref{tab41}. The collected results in Table \ref{tab41} denote that kNN outperforms the SVM and RF in terms of average accuracy and f-measure. Furthermore, it is evident that moving from dataset $\mathcal{D}_1$ to $\mathcal{D}_6$ the attained values are increased because of the fact that the level of noise is decreased from 10 dB to 70 dB. 

\begin{table}
\centering
\caption{The attained accuracy and f-measure values by means of the baseline models.}\label{tab41}
\begin{tabular}{lllllll}
\hline\hline Dataset & \multicolumn{2}{c}{kNN} & \multicolumn{2}{c}{SVM} & \multicolumn{2}{c}{RF}\\
\hline & $\mathcal{A}$ & $\mathcal{F}_1$ & $\mathcal{A}$ & $\mathcal{F}_1$ & $\mathcal{A}$ & $\mathcal{F}_1$\\
\hline $\mathcal{D}_1$ & 0.4288 & 0.4413 & 0.3559 & 0.3717 & 0.4012 & 0.4014\\
$\mathcal{D}_2$ & 0.4257 & 0.4424 & 0.3480 & 0.3651 & 0.1793 & 0.1797\\
$\mathcal{D}_3$ & 0.5716 & 0.5807 & 0.3514 & 0.3676 & 0.3417 & 0.3452\\
$\mathcal{D}_4$ & 0.5786 & 0.5858 & 0.5531 & 0.5692 & 0.3403 & 0.3429\\
$\mathcal{D}_5$ & 0.6778 & 0.6848 & 0.7748 & 0.7887 & 0.5037 & 0.5053\\
$\mathcal{D}_6$ & 0.6804 & 0.6869 & 0.7211 & 0.7370 & 0.5150 & 0.5176\\
\hline Avg. & 0.5604 & 0.5703 & 0.5174 & 0.5332 & 0.3802 & 0.3820 \\
\hline\hline
\end{tabular}
\end{table}

\subsection{FS-classifier Combinations}\label{sec42}
In this section, the performance of the FS techniques along with their combinations with the given classification models is presented in terms of the accuracy and f-measure. The aim is to first compare the performance of the FS techniques to see which technique is superior, which is then followed by a discussion on the superiority of FS-classifier combinations to see which combination could be the best for the sake of fault diagnosis in power systems. 

\begin{table*}
\centering
\setlength{\tabcolsep}{2pt}
\scriptsize
\caption{The attained average values of the accuracy and f-measure for FS-classifier combinations.}\label{tab421}
\begin{tabular}{l|cccccc|cccccc|cccccc|cccccc|cccccc}
\hline\hline \multirow{2}{*}{Dataset} & \multicolumn{6}{c}{InfFS} & \multicolumn{6}{c}{relief} & \multicolumn{6}{c}{LASSO} & \multicolumn{6}{c}{UFSOL} & \multicolumn{6}{c}{CFMI}\\ 
\cline{2-31} & \multicolumn{2}{c}{kNN} & \multicolumn{2}{c}{SVM} & \multicolumn{2}{c}{RF} & \multicolumn{2}{c}{kNN} & \multicolumn{2}{c}{SVM} & \multicolumn{2}{c}{RF} & \multicolumn{2}{c}{kNN} & \multicolumn{2}{c}{SVM} & \multicolumn{2}{c}{RF} & \multicolumn{2}{c}{kNN} & \multicolumn{2}{c}{SVM} & \multicolumn{2}{c}{RF} & \multicolumn{2}{c}{kNN} & \multicolumn{2}{c}{SVM} & \multicolumn{2}{c}{RF}\\
\hline $\mathcal{D}_1$ & 0.54 & 0.51 & 0.60 & 0.58 & 0.56 & 0.51 & 0.58 & 0.59 & 0.65 & 0.67 & 0.60 & 0.57 & 0.58 & 0.54 & 0.66 & 0.64 & 0.60 & 0.55 & 0.56 & 0.42 & 0.62 & 0.55 & 0.58 & 0.58 & 0.59 & 0.58 & 0.73 & 0.74 & 0.56 & 0.56 \\
$\mathcal{D}_2$ & 0.57 & 0.56 & 0.64 & 0.64 & 0.60 & 0.56 & 0.61 & 0.56 & 0.68 & 0.65 & 0.63 & 0.56 & 0.61 & 0.56 & 0.67 & 0.64 & 0.62 & 0.56 & 0.57 & 0.47 & 0.64 & 0.57 & 0.60 & 0.60 & 0.59 & 0.57 & 0.73 & 0.73 & 0.56 & 0.55  \\
$\mathcal{D}_3$ & 0.64 & 0.68 & 0.69 & 0.73 & 0.64 & 0.67 & 0.74 & 0.77 & 0.79 & 0.82 & 0.74 & 0.76 & 0.70 & 0.73 & 0.75 & 0.78 & 0.71 & 0.72 & 0.66 & 0.60 & 0.70 & 0.65 & 0.66 & 0.59 & 0.71 & 0.71 & 0.78 & 0.79 & 0.62 & 0.61 \\
$\mathcal{D}_4$ & 0.71 & 0.74 & 0.76 & 0.79 & 0.71 & 0.73 & 0.73 & 0.75 & 0.78 & 0.81 & 0.73 & 0.75 & 0.72 & 0.74 & 0.77 & 0.79 & 0.73 & 0.74 & 0.66 & 0.61 & 0.71 & 0.66 & 0.67 & 0.61 & 0.73 & 0.73 & 0.79 & 0.79 & 0.64 & 0.63 \\
$\mathcal{D}_5$ & 0.80 & 0.83 & 0.83 & 0.86 & 0.79 & 0.82 & 0.74 & 0.79 & 0.78 & 0.82 & 0.75 & 0.79 & 0.76 & 0.75 & 0.80 & 0.79 & 0.78 & 0.77 & 0.75 & 0.73 & 0.78 & 0.77 & 0.74 & 0.71 & 0.82 & 0.81 & 0.82 & 0.83 & 0.71 & 0.70\\
$\mathcal{D}_6$ & 0.75 & 0.79 & 0.79 & 0.82 & 0.75 & 0.78 & 0.77 & 0.81 & 0.81 & 0.84 & 0.78 & 0.82 & 0.70 & 0.75 & 0.73 & 0.77 & 0.72 & 0.76 & 0.77 & 0.74 & 0.80 & 0.77 & 0.76 & 0.70 & 0.82 & 0.82 & 0.83 & 0.84 & 0.76 & 0.75 \\
\hline Avg. & 0.67 & 0.69 & 0.72 & 0.74 & 0.68 & 0.68 & 0.70 & 0.71 & 0.75 & 0.77 & 0.71 & 0.71 & 0.68 & 0.68 & 0.73 & 0.73 & 0.70 & 0.68 & 0.66 & 0.60 & 0.71 & 0.66 & 0.67 & 0.63 & 0.71 & 0.70 & 0.78 & 0.79 & 0.64 & 0.63 \\
\hline\hline
\end{tabular}
\end{table*}

We begin with a comprehensive study on the performance of the given FS-classifier combinations, where the attained results are collected in Table \ref{tab421} in terms of accuracy and f-measure. We initially study the effects of the SNR values on the performance of combinations. From the presented results in Table \ref{tab421}, it could be concluded that the average values of accuracy and f-measure increases from 0.61, 0.71, 0.77 and 0.58, 0.72, 0.78, respectively, by moving from SNR values of 10 dB to 30 dB and 70 dB. Obviously, the lower the level of noise, the better the performance of the combinations. In order to check for the performance of the given combinations in dealing with different fault resistance values, by resorting to the collected results in Table \ref{tab421} it could be concluded that the average values of the accuracy and f-measure are 0.69, 0.69 and 0.70, 0.70, when the fault resistance values are 10$\Omega$ and 1$\Omega$, respectively. Therefore, the attained results denote that the value of the fault resistance has slightly impacted the performance of the given combinations, where it shows that a higher value of the fault resistance could negatively impact the performance. 

\begin{table}
\centering
\setlength{\tabcolsep}{10pt}
\caption{Comparison between various FS techniques in terms of The average accuracy and f-measure for the given datasets.}\label{tab422}
\begin{tabular}{lccccc}
\hline\hline FS & InfFS & relief & LASSO & UFSOL & CFMI\\
\hline $\mathcal{A}$ & 0.69 & 0.71 & 0.70 & 0.68 & 0.72\\
$\mathcal{F}_1$ & 0.70 & 0.72 & 0.70 & 0.63 & 0.71\\
\hline\hline
\end{tabular}
\end{table}

In terms of the comparison between the given FS techniques, the average values of the accuracy and f-measure w.r.t. to datasets $\mathcal{D}_1$ to $\mathcal{D}_6$ are summarized in Table \ref{tab422}. Firstly, it could be observed that CFMI outperforms the other FS techniques in terms of accuracy, which is then followed by relief, LASSO, InfFS, and UFSOL. However, in terms of the f-measure, the collected results denote that the relief technique outperforms the rest of the techniques, which is followed by CFMI, LASSO, InfFS, and UFSOL. Secondly, by taking into account the average values of the accuracy for the filter, wrapper, and embedded techniques, the attained results denote that all categories have almost the same performance, however, in terms of the f-measure, the filter category of methods outperforms the other two categories by 1\%. 

\begin{table}
\centering
\setlength{\tabcolsep}{20pt}
\caption{The attained average values of the accuracy and f-measure by means of the classification models in combination with the given FS techniques.}\label{tab423}
\begin{tabular}{lccc}
\hline\hline Model & kNN & SVM & RF\\
\hline $\mathcal{A}$ & 0.68 & 0.74 & 0.68 \\
$\mathcal{F}_1$ & 0.67 & 0.73 & 0.67\\
\hline\hline
\end{tabular}
\end{table}

In terms of the classification models, i.e., kNN, SVM, and RF, the average values of the accuracy and f-measure w.r.t. all combinations are summarized in Table \ref{tab423}. The collected results denote that SVM outperforms the other classification models, while kNN and RF have shown the same performance in dealing with the given datasets and in accordance with the constructed combinations. Further to this, by comparing the results of Table \ref{tab423} with those presented in Table \ref{tab41}, it is evident that FS techniques have considerably helped with improving the performance of the given classification models. Turning into details, it could be observed that the performance of the kNN is improved from 0.56 to 0.68, that is from 0.51 to 0.74 and from 0.38 to 0.68 for the SVM and RF, respectively, and in terms of the average accuracy values. In the same vein and in terms of the average f-measure values, the presented results show that the performance of the kNN is improved from 0.57 to 0.67, that is from 0.53 to 0.73 and from 0.38 to 0.67 for the SVM and RF, respectively. 

\begin{table}
\centering
\setlength{\tabcolsep}{20pt}
\caption{Ranking of the DR-classifier combinations based on the accuracy and f-measure.}\label{tab424}
\begin{tabular}{lll}
\hline\hline Rank & Combination & Combination\\
\hline & Accuracy & f-measure\\
\hline 1 & CFMI-SVM & CFMI-SVM\\
2 & relief-SVM & relief-SVM \\
3 & LASSO-SVM & InfFS-SVM \\
4 & InfFS-SVM & LASSO-SVM \\
5 & CFMI-kNN & relief-kNN \\
6 & UFSOL-SVM & relief-RF \\
7 & relief-RF & CFMI-kNN \\
8 & relief-kNN & InfFS-kNN \\
9 & LASSO-RF & InfFS-RF \\
10 & LASSO-kNN & LASSO-kNN \\
11 & InfFS-RF & LASSO-RF \\
12 & InfFS-kNN & UFSOL-SVM \\
13 & UFSOL-RF & CFMI-RF \\
14 & UFSOL-kNN & UFSOL-RF \\
15 & CFMI-RF & UFSOL-kNN \\
\hline\hline
\end{tabular}
\end{table}

Finally, we analyze the performance of the constructed combinations in order to rank them based on the accuracy or f-measure. In this regard, the ranked combinations are listed in Table \ref{tab424} w.r.t. accuracy and f-measure. Firstly, the given ranking in terms of the accuracy and f-measure denotes that CFMI-SVM is the best combination in order to deal with the constructed datasets $\mathcal{D}_1$ to $\mathcal{D}_6$. Secondly, it could be observed that in four out of the five first combinations, the FS techniques are combined with the SVM, showing that SVM could be of a better performance in combination with FS techniques for the sake of fault diagnosis. 

\subsection{DR-classifier Combinations}\label{sec43}
In this section, we present the attained results by means of the DR-classifier combinations. In this regard, we first compare DR techniques in dealing with the given datasets in terms of accuracy and f-measure, and, then, all the possible DR-classifier combinations are ranked based on their performance.

Same as what was presented in Section \ref{sec42} for the FS-classifier combinations, Table \ref{tab431} summarizes the attained average values of the accuracy and f-measure of all DR-classifier combinations w.r.t. datasets $\mathcal{D}_1$ to $\mathcal{D}_6$. We start with the impact of the SNR on the performance of the given DR techniques. In this regard, we resort to the attained accuracy and f-measure values collected in Table \ref{tab431} and calculate the average accuracy and f-measure for all combinations w.r.t. different SNR values. To this end, the attained average accuracy and f-measures are 0.55, 0.62, 0.68, 0.57, 0.64, and 0.69, respectively, for datasets with the SNR value of 10 dB, 30 dB, and 70 dB. As it was expected, the lower the value of the SNR, the lower the value of accuracy and f-measure values, denoting that noisy measurements could degrade the performance of DR techniques. Other than the SNR, the value of the fault resistance could also play an important role in the performance of the given DR techniques. By resorting to the results presented in Table \ref{tab431}, the average values of the accuracy and f-measure w.r.t. fault resistance values of 1$\Omega$ and 10$\Omega$ are 0.62, 0.61, and 0.64, 0.63, respectively, denoting that both the index measures are degraded by 1\% when the fault resistance value is increased from 1$\Omega$ to 10$\Omega$. This is due to the fact that a higher fault resistance value could decrease the signature of the faults on the collected data measurements from the grid. 

\begin{table*}
\centering
\setlength{\tabcolsep}{2pt}
\scriptsize
\caption{The attained average values of the accuracy and f-measure for DR-classifier combinations.}\label{tab431}
\begin{tabular}{l|cccccc|cccccc|cccccc|cccccc|cccccc}
\hline\hline \multirow{2}{*}{Dataset} & \multicolumn{6}{c}{PCA} & \multicolumn{6}{c}{MDS} & \multicolumn{6}{c}{LDA} & \multicolumn{6}{c}{LLE} & \multicolumn{6}{c}{CADR}\\ 
\cline{2-31} & \multicolumn{2}{c}{kNN} & \multicolumn{2}{c}{SVM} & \multicolumn{2}{c}{RF} & \multicolumn{2}{c}{kNN} & \multicolumn{2}{c}{SVM} & \multicolumn{2}{c}{RF} & \multicolumn{2}{c}{kNN} & \multicolumn{2}{c}{SVM} & \multicolumn{2}{c}{RF} & \multicolumn{2}{c}{kNN} & \multicolumn{2}{c}{SVM} & \multicolumn{2}{c}{RF} & \multicolumn{2}{c}{kNN} & \multicolumn{2}{c}{SVM} & \multicolumn{2}{c}{RF}\\
\hline $\mathcal{D}_1$ & 0.49 & 0.50 & 0.63 & 0.64 & 0.58 & 0.58 & 0.50 & 0.50 & 0.62 & 0.63 & 0.58 & 0.57 & 0.40 & 0.50 & 0.43 & 0.51 & 0.42 & 0.50 & 0.41 & 0.42 & 0.58 & 0.65 & 0.46 & 0.46 & 0.73 & 0.74 & 0.81 & 0.81 & 0.72 & 0.72\\
$\mathcal{D}_2$ & 0.48 & 0.48 & 0.64 & 0.64 & 0.57 & 0.57 & 0.48 & 0.48 & 0.62 & 0.62 & 0.57 & 0.57 & 0.39 & 0.50 & 0.42 & 0.50 & 0.42 & 0.50 & 0.43 & 0.43 & 0.54 & 0.45 & 0.46 & 0.46 & 0.73 & 0.73 & 0.80 & 0.80 & 0.70 & 0.70\\
$\mathcal{D}_3$ & 0.60 & 0.61 & 0.71 & 0.71 & 0.67 & 0.67 & 0.60 & 0.60 & 0.71 & 0.71 & 0.67 & 0.67 & 0.39 & 0.50 & 0.43 & 0.51 & 0.42 & 0.50 & 0.51 & 0.52 & 0.70 & 0.75 & 0.56 & 0.56 & 0.76 & 0.77 & 0.85 & 0.86 & 0.74 & 0.74\\
$\mathcal{D}_4$ & 0.59 & 0.59 & 0.70 & 0.71 & 0.66 & 0.66 & 0.59 & 0.59 & 0.70 & 0.71 & 0.66 & 0.66 & 0.40 & 0.51 & 0.43 & 0.51 & 0.42 & 0.50 & 0.50 & 0.51 & 0.67 & 0.73 & 0.55 & 0.56 & 0.77 & 0.77 & 0.86 & 0.87 & 0.74 & 0.75\\
$\mathcal{D}_5$ & 0.68 & 0.68 & 0.78 & 0.78 & 0.74 & 0.74 & 0.68 & 0.68 & 0.78 & 0.79 & 0.74 & 0.74 & 0.39 & 0.50 & 0.43 & 0.50 & 0.42 & 0.50 & 0.56 & 0.56 & 0.87 & 0.67 & 0.62 & 0.62 & 0.82 & 0.82 & 0.90 & 0.91 & 0.79 & 0.79\\
$\mathcal{D}_6$ & 0.67 & 0.67 & 0.77 & 0.78 & 0.73 & 0.73 & 0.67 & 0.67 & 0.77 & 0.78 & 0.73 & 0.73 & 0.39 & 0.50 & 0.43 & 0.50 & 0.43 & 0.51 & 0.57 & 0.57 & 0.83 & 0.85 & 0.62 & 0.62 & 0.83 & 0.83 & 0.90 & 0.91 & 0.80 & 0.80\\
\hline Avg. & 0.59 & 0.59 & 0.71 & 0.71 & 0.66 & 0.66 & 0.59 & 0.59 & 0.70 & 0.71 & 0.66 & 0.66 & 0.39 & 0.50 & 0.43 & 0.51 & 0.42 & 0.50 & 0.50 & 0.50 & 0.69 & 0.68 & 0.55 & 0.55 & 0.77 & 0.78 & 0.85 & 0.86 & 0.75 & 0.75\\
\hline\hline
\end{tabular}
\end{table*}

In terms of the comparison between the given DR techniques, Table \ref{tab432} summarizes the performance of all the given techniques in terms of the average accuracy and f-measure w.r.t. datasets $\mathcal{D}_1$ to $\mathcal{D}_6$. Following the presented results in Table \ref{tab432}, it could be observed that firstly, CADR outperforms other DR techniques in terms of accuracy and f-measure. Secondly, for the linear DR techniques, i.e., PCA and LDA, the average accuracy and f-measure values are 0.53 and 0.58, respectively, while for the nonlinear techniques, i.e., MDS, LLE, and CADR, the average accuracy, and f-measure values are 0.67 and 0.67, respectively. Therefore, the attained results denote that nonlinear DR techniques could in general outperform linear ones for the sake of fault diagnosis. 

\begin{table}
\centering
\setlength{\tabcolsep}{10pt}
\caption{Comparison between various DR techniques in terms of The average accuracy and f-measure for the given datasets.}\label{tab432}
\begin{tabular}{lccccc}
\hline\hline DR & PCA & MDS & LDA & LLE & CADR\\
\hline $\mathcal{A}$ & 0.65 & 0.65 & 0.41 & 0.58 & 0.79\\
$\mathcal{F}_1$ & 0.65 & 0.65 & 0.50 & 0.58 & 0.80\\
\hline\hline
\end{tabular}
\end{table}

In terms of the classification models combined with the given DR techniques, the average accuracy and f-measure values are summarized in Table \ref{tab433}. Firstly, it is evident that SVM results in the best classification performance ($\mathcal{A}=0.68$, $\mathcal{F}_1=0.69$), which is followed by the RF ($\mathcal{A}=0.61$, $\mathcal{F}_1=0.62$) and kNN ($\mathcal{A}=0.57$, $\mathcal{F}_1=0.59$). Secondly, compared with the results presented in Table \ref{tab41} for the baseline models, it could be concluded that the constructed DR-classifier combinations have improved the performance of the diagnostic system, verifying the importance of feature extraction in dealing with the curse of dimensionality. 

\begin{table}
\centering
\setlength{\tabcolsep}{20pt}
\caption{The attained average values of the accuracy and f-measure by means of the classification models in combination with the given DR techniques.}\label{tab433}
\begin{tabular}{lccc}
\hline\hline Model & kNN & SVM & RF\\
\hline $\mathcal{A}$ & 0.57 & 0.68 & 0.61\\
$\mathcal{F}_1$ & 0.59 & 0.69 & 0.62\\
\hline\hline
\end{tabular}
\end{table}

Finally, all the possible DR-classifier combinations are ranked based on their performance in terms of accuracy and f-measure, where the ranked combinations are collected in Table \ref{tab434}. The ranked combinations in terms of accuracy, in line with what was discussed previously, denote that CADR in combination with any of the given classification models outperforms the rest of the DR techniques. In addition, three out of the first five combinations are cases in which SVM is combined with a DR technique, verifying the superiority of the SVM classification model over the kNN and RF. Furthermore, seven out of the first 10 combinations include a nonlinear DR technique that denotes these techniques are superior to their linear counterparts. The provided rankings based upon the f-measure index, there is no change in the first 10 combinations, and, therefore, the same concluding remarks could be made. 

\begin{table}
\centering
\setlength{\tabcolsep}{20pt}
\caption{Ranking of the DR-classifier combinations based on the accuracy and f-measure.}\label{tab434}
\begin{tabular}{lll}
\hline\hline Rank & Combination & Combination\\
\hline & Accuracy & f-measure\\
\hline 1 & CADR-SVM & CADR-SVM\\
2 & CADR-kNN & CADR-kNN\\
3 & CADR-RF & CADR-RF\\
4 & PCA-SVM & PCA-SVM\\
5 & MDS-SVM & MDS-SVM\\
6 & LLE-SVM & LLE-SVM\\
7 & PCA-RF & PCA-RF\\
8 & MDS-RF & MDS-RF\\
9 & PCA-kNN & PCA-kNN \\
10 & MDS-kNN & MDS-kNN\\
11 & LLE-RF & LLE-RF\\
12 & LLE-kNN & LDA-SVM\\
13 & LDA-SVM & LDA-kNN\\
14 & LDA-RF & LLE-kNN\\
15 & LDA-kNN & LDA-RF\\
\hline\hline
\end{tabular}
\end{table}

\subsection{Discussion}\label{sec44}
In Section \ref{sec42} and Section \ref{sec43}, we analyzed the performance of the FS/DR-classifier combinations. In this section, we aim to discuss and compare the performance of FS techniques in comparison with the DR techniques in order to find out which type of data reduction could help more with improving the performance of fault diagnosis for the given case-study. 

In this regard, we resort to the average values of the accuracy and f-measure attained by means of FS techniques in combination with all the given classification models and compare them with those of DR techniques. The attained results denote that the average accuracy and f-measure values for the FS techniques are 0.70 and 0.70, respectively, and those are 0.62 and 0.64 for the DR techniques. Therefore, the attained results denote that for this case-study, the combination of FS techniques with classification models could deal better with the classification task compared with the combinations of the DR techniques with classification models. 

\section{Concluding Remarks}\label{sec5}
In this paper, we studied the fault diagnosis problem of cyber-physical power systems by resorting to a data-driven technique. This proposal suggested the combination of data reduction techniques including feature selection and dimensionality reduction and their combinations with classification models in order to identify four types of faults including generator outage, generator ground, and load loss. It was proposed to make use of different types of feature selection and dimensionality reduction techniques to investigate the impact of different models on diagnostic performance. The attained results first denoted that in terms of feature selection and their combinations with classification models, all the categories performed more or less the same to deal with the classification task. However, for the dimensionality reduction techniques, it was observed that nonlinear models could improve the classification performance in comparison with the linear counterpart. Furthermore, the results of the given experiments denoted that in general, feature selection techniques could be of a better performance in comparison with dimensionality reduction methods for the sake of fault diagnosis. In addition to this, we studied the impact of noisy measurements and fault resistance values, where the results showed that when the level of noise and the value of the fault resistance decreases, the performance of the constructed combinations could be improved. Due to the high volume of available data that could be collected from large-scale power systems, this study could be further extended to the use of deep learning algorithms for the sake of classification due to their ability in dealing with large-size datasets. 




\end{document}